\documentclass[conference]{IEEEtran}
\IEEEoverridecommandlockouts
\usepackage{cite}
\usepackage{amsmath,amssymb,amsfonts}
\usepackage{graphicx}
\usepackage{textcomp}
\usepackage{xcolor}
\usepackage{booktabs}
\usepackage{multirow}
\usepackage{array}
\usepackage{stfloats}
\usepackage{pifont}   
\usepackage{microtype}   
\usepackage{enumitem}    

\usepackage[colorlinks=true,allcolors=blue!55!black]{hyperref}

\begin{document}

\title{Rolling Day-Wise Mortality Prediction in Critically Ill Patients With AKI on CRRT Utilizing Machine Pressure Waveforms}

\author{
\IEEEauthorblockN{
Shehan Irteza Pranto\textsuperscript{1},
Joanna Yang\textsuperscript{6},
Joshua Lambert\textsuperscript{3},
Stuart L. Goldstein\textsuperscript{4},\\
Lili Chan\textsuperscript{5,6},
Girish N. Nadkarni\textsuperscript{5,6},
Tiago K. Colicchio\textsuperscript{7},
Javier A. Neyra\textsuperscript{2},
Jin Chen\textsuperscript{2,}\textsuperscript{*}
\thanks{\textsuperscript{*}Corresponding author}
}
\IEEEauthorblockA{\footnotesize
\textsuperscript{1}\textit{Department of Biomedical Informatics and Data Science, University of Alabama at Birmingham, AL, USA}\\
\textsuperscript{2}\textit{Division of Nephrology, Department of Medicine, University of Alabama at Birmingham, AL, USA}\\
\textsuperscript{3}\textit{College of Nursing, University of Cincinnati, OH, USA}\quad
\textsuperscript{4}\textit{Department of Pediatrics, University of Cincinnati, OH, USA}\\
\textsuperscript{5}\textit{Department of Medicine, Division of Nephrology at the Icahn School of Medicine at Mount Sinai, New York, NY, USA}\\
\textsuperscript{6}\textit{Windreich Department of Artificial Intelligence and Human Health at the Icahn School of Medicine at Mount Sinai, New York, NY, USA}\\
\textsuperscript{7}\textit{Department of Health Economics and Systems Policy, Peter O’Donnell Jr. School of Public Health UT Southwestern medical center, Dallas, TX, USA}
}
}

\maketitle

\begin{abstract}
Critically ill patients with acute kidney injury (AKI) on continuous renal replacement therapy (CRRT) face high mortality, yet current risk assessment methods rely primarily on clinical parameters from electronic health records (EHR) and ignore the rich, high-frequency minute-level circuit pressure waveforms generated by CRRT machines that dynamically track the extracorporeal circuit's interaction with the patient. Clinicians therefore cannot see deterioration as it develops. As a result, risk is reassessed only when labs are drawn, while the one continuous record of the patient--circuit interaction is discarded because, as logged (with minute-level frequency), it is uninterpretable, contaminated by shared-device records, non-physiological minutes, and sensor artifact. To make the stream usable, we aligned machine records to charted therapy intervals (preventing cross-patient leakage), remove non-physiological priming and downtime minutes, tune denoising on a synthetic spike-injection benchmark, and mask unobserved intervals rather than imputing them. On this cleaned stream we define a rolling day-wise task and a transformer-based stacked ensemble that late-fuses a window-reduced sequence transformer with classical models over engineered circuit-instability features and clinical EHR variables. In a leak-safe benchmark on the multi-center CRRTnet cohort (976 patients, 4,585 treatment days), the machine-only model carries the lowest standalone prognostic value (AUROC 0.625), followed my the EHR-only model (0.717). Integrating clinical EHR and machine streams reaches a one-day mortality AUROC of 0.766, above both standalone models. SHAP attribution further showed that explicit circuit-instability descriptors raised the machine share of the top-15 combined-model features from 3 to 7 (20.0\% to 46.7\%), localized to interpretable mechanical dynamics: filter pressure, transmembrane pressure (TMP), and access-to-return difference (ARD). To our knowledge, this is the first patient-level mortality prediction incorporating CRRT machine data, turning a discarded bedside stream into a continuous risk signal.
\end{abstract}

\begin{IEEEkeywords}
mortality prediction, acute kidney injury, CRRT, clinical time series, signal preprocessing, machine learning
\end{IEEEkeywords}

\section{Introduction}

Acute kidney injury (AKI) requiring continuous renal replacement therapy (CRRT) is a frequent and serious complication in critically ill patients, carrying in-hospital mortality exceeding 50\%~\cite{russo2019, negi2016, negi2023}. Safe bedside management, such as determining timely therapy adjustments or recognizing rapid clinical decompensation, requires accurate, continuously updated short-horizon risk estimates~\cite{patel2022, notaro2026}. This need has motivated growing interest in dynamic clinical decision support (CDS) systems. However, most existing approaches rely on intermittently sampled electronic health record (EHR) data, which provide delayed and static snapshots that are insufficient to capture rapidly evolving physiology~\cite{deasy2020, choi2022}.

Simultaneously, a major untapped opportunity exists in the continuous, high-resolution machine data passively generated by CRRT devices. This machine provide minute-level waveforms of four distinct circuit pressures (access, filter, effluent, return) and their derived gradients dynamically track the extracorporeal circuit's physical state, providing early indicators of clotting, membrane fouling/clogging, and downstream resistance~\cite{sansom2019}. We hypothesize that these continuous mechanical fluctuations defining circuit instability inherently encode real-time information about the patient's underlying physiological trajectory; however, this dense signal is currently discarded or used only for monitoring mechanical endpoints~\cite{yang2024}, leaving its prognostic value for patient outcomes entirely unexplored.

Consequently, existing machine learning approaches for AKI-CRRT outcomes rely almost exclusively on structured EHR data, targeting fixed horizons such as in-hospital, 28-day, or 90-day mortality~\cite{kang2020,gu2024,thadani2024}. State-of-the-art models for mortality prediction, CRRT liberation~\cite{zhong2024,zhu2024}, or renal recovery~\cite{zamanzadeh2024} typically employ random forests, gradient boosting such as XGBoost and LightGBM, or logistic regression using variables such as SOFA/APACHE-II scores, vital signs, and laboratory measurements. More advanced approaches incorporate temporal dynamics through recurrent neural networks (RNNs)~\cite{liu2022kit}, or augment structured EHR data with unstructured clinical notes~\cite{zha2022}. Both still rely on measurements recorded at irregular, clinically driven intervals, so physiologic changes occurring between observations are not captured. Conversely, when high-resolution CRRT machine data have been used, they are typically restricted to circuit-level tasks such as predicting filter clotting or circuit lifespan~\cite{yang2024,buccione2025}, often using LSTM-based models on transmembrane pressure (TMP), the pressure gradient across the hemofilter membrane that rises as the filter becomes fouled~\cite{wang2025tmp} or tree-based classifiers for coagulation detection, that is, identifying clot formation in the circuit as it occurs rather than predicting it in advance~\cite{buccione2025}. Thus, despite their rich physiological content, continuous CRRT waveforms have not been leveraged for patient-level mortality prediction in a rolling, day-wise framework.

This stream has gone unused because of it's complexity in modeling. Devices are
moved between patients while logging continuously, so naive extraction assigns
one patient's minutes to another; priming and downtime minutes are mechanically
valid but carry no physiological information; sensor artifact dominates unfiltered variance statistics; and interpolating across genuinely unobserved intervals fabricates trends that a model will learn. Each failure inflates rather than degrades apparent performance, so establishing that the waveform is prognostic requires first establishing that the extracted waveform is real.

To bridge this gap, we propose a rolling day-wise mortality prediction framework that independently scores each treatment day from its own minute-level CRRT signal. Our contributions are threefold:

\begin{itemize}[leftmargin=1.1em,topsep=2pt,itemsep=2pt,parsep=0pt,partopsep=0pt]
\item \textbf{A reproducible pipeline for raw CRRT machine data.} Machine
timestamps are aligned to charted therapy intervals to eliminate
device-sharing leakage (removing 13.5\% of raw minutes), priming and downtime
minutes are excised by run-time and effluent-increment rules (a further
1.8\%), the denoising operating point is fixed on a synthetic spike-injection
benchmark, and short dropouts are interpolated while longer gaps stay masked.

\item \textbf{First evidence that CRRT waveforms carry patient-level prognostic signal.} Minute-level circuit pressure alone predicts next-day mortality above chance (AUROC 0.625) with no clinical data, and adding it to routine EHR features raises discrimination from a 0.717 clinical-only reference to 0.766, so the signal is not redundant with the labs.

\item \textbf{A leak-safe rolling formulation and two-level stacked ensemble.} We define a causal day-wise task with an excluded gap window, late-fuse a Transformer encoder~\cite{vaswani2017} over a window-reduced representation of the pressure stream with classical models trained on engineered circuit-instability features and clinical EHR variables, and report the full protocol with all preprocessing, model selection, stacking weights, and thresholds fit on train or
validation only, leaving the test split untouched.
\end{itemize}

\begin{figure}[!t]
\centering
\includegraphics[
    width=\columnwidth,
    trim={0.6cm 0.4cm 1.3cm 0.2cm},
    clip
]{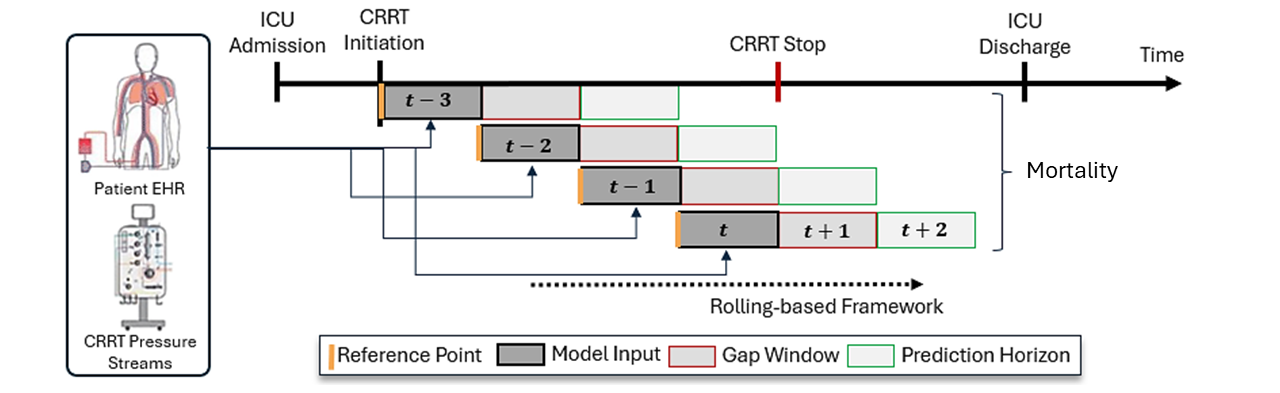}
\caption{Rolling day-wise prediction framework. The window advances one day at a time ($t{-}3,\dots,t$). Each treatment day defines a model-input window (dark), separated from the prediction horizon (green) by an excluded gap window (red) to guarantee causal, prospective risk estimates.}
\label{fig:rolling}
\end{figure}

\section{Cohort and Prediction Task}

\subsection{Cohort}
This study utilizes CRRTnet~\cite{heung2017,rewa2023}, a large multi-institutional registry of adults with AKI on CRRT (2013--2021) that provides linked EHR clinical data with machine data. Restricting the analysis to the first seven treatment days (with an average of days on CRRT of 5.32), the evaluated cohort comprises 976 patients, 1,153 continuous treatment episodes, and 4,585 treatment days. At approximately 1,440 timestamps per
treatment day, each day is a long, high-volume sequence rather
than a handful of charted values. The cohort is partitioned at the patient level using a stratified 70:10:20 split: a training set (682 patients, 3,236 days, 6.2\% prevalence), a validation set (98 patients, 462 days, 6.5\% prevalence), and a held-out test set (196 patients, 887 days, 7.0\% prevalence). Patient-level baseline characteristics at CRRT initiation, stratified by in-hospital mortality (53.8\% overall), are reported in Table~\ref{tab:baseline}. However, the primary modeling target is the treatment-day mortality label.

\subsection{Rolling day-wise labeling}
The prediction task is formulated as a rolling, day-wise scheme (Fig.~\ref{fig:rolling}). Within a treatment episode, each treatment day forms one model-input window comprising that day's CRRT pressure stream together with the patient's EHR data available up to that day; baseline covariates fixed at CRRT initiation plus forward-filled laboratory values, with no look-ahead. A one-day gap immediately following the input is intentionally excluded as a clinical-intervention buffer, ensuring strictly causal evaluation. The window advances one day at a time to yield independent, continually updating risk estimates; an input day is labeled positive if the patient dies on the day following the gap (the one-day horizon).

\begin{table}[!tb]
\caption{Patient baseline characteristics by in-hospital mortality.}
\label{tab:baseline}
\centering
\scriptsize
\setlength{\tabcolsep}{3.5pt}
\begin{tabular}{lccc}
\toprule
Variable & Survivors ($n{=}451$) & Non-surv. ($n{=}525$) & $p$ \\
\midrule
Age (Years) & 57.0 [48.0--67.0] & 63.0 [55.0--70.2] & $<$.001 \\
Sex (Male) & 205 (58.7\%) & 238 (63.3\%) & .237 \\
BMI, kg/m$^2$ & 30.1 [25.2--36.6] & 29.7 [25.2--36.1] & .944 \\
Charlson score & 2.0 [1.0--3.0] & 2.0 [1.0--4.0] & $<$.001 \\
APACHE-II (part.) & 9.0 [5.0--25.0] & 13.0 [5.0--25.0] & .037 \\
\bottomrule
\end{tabular}

\vspace{1pt}
\footnotesize{\textit{Note:} Continuous variables are presented as median [IQR] with Mann--Whitney $p$-values; binary variables as $n$ (\%) with $\chi^2$ $p$-values.}
\end{table}

\section{Method}

\subsection{Signal acquisition and cleaning}
\label{sec:cleaning}
\begin{figure*}[!t]
\centering
\includegraphics[width=0.95\textwidth]{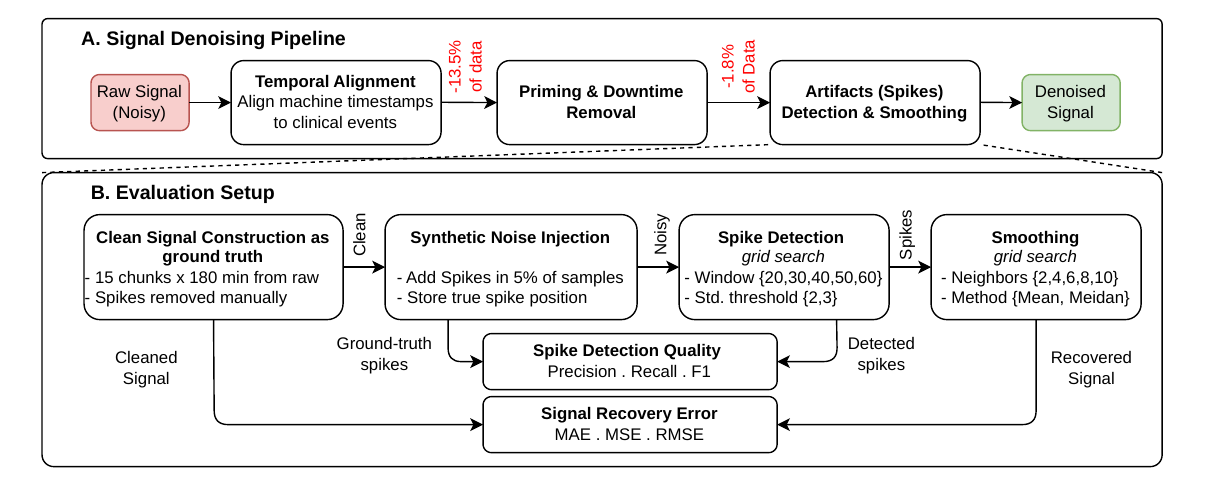}
\caption{Signal-processing pipeline (top) and evaluation (bottom). The raw pressure stream is temporally aligned to clinical events, stripped of priming and downtime minutes, and denoised. The denoising operating point is optimized on a synthetic spike-injection benchmark (5\% corruption): grid searches over
detection parameters (window length, standard-score threshold) and smoothing configurations (neighbor count, mean/median aggregation) maximize spike-localization $F_1$ and minimize reconstruction error (MAE, MSE).}
\label{fig:flow}
\end{figure*}

The continuous input stream comprises minute-level circuit pressures (access $A$, filter $F$, effluent $E$, return $R$), alongside machine run-time ($RT$) and cumulative effluent ($E$) counters ~\cite{tandukar2019, macedo2016}. To produce the denoised signal (Fig.~\ref{fig:flow}A), the raw data undergoes three primary cleaning steps:

\textbf{Temporal alignment.} CRRT devices are shared across patients and continue logging between therapies, so machine records are attributed to a patient only within that patient's charted therapy interval. Machine timestamps ($t_i$) are synchronized with EHR CRRT start ($t_{\mathrm{start}}$) and stop ($t_{\mathrm{stop}}$) times, and out-of-bounds minutes are dropped using a 12-hour buffer $(\tau_{\mathrm{buffer}})$ that absorbs delayed clinical charting without admitting a neighboring patient's session:
\begin{equation}
\mathrm{drop}_i = (t_i < t_{\mathrm{start}})\vee(t_i > t_{\mathrm{stop}}+\tau_{\mathrm{buffer}}).
\end{equation}
This alignment step removed 13.5\% of the raw machine data.

\textbf{Priming and downtime removal.} Non-physiological minutes from initial circuit priming (identified by zero run-time) and transient machine downtimes (identified by a zero effluent increment)~\cite{kdigo2012} reflect the operating state of the device rather than the state of the patient, and they generate long low-variance runs that bias any distributional summary of the day. They are excluded by:
\begin{equation}
\mathrm{drop}_i = (\Delta t_i{=}1)\wedge(\mathrm{RT}_i{=}0)\wedge(\mathrm{RT}_{i-1}{=}0),
\end{equation}
\begin{equation}
\mathrm{drop}_i = (\Delta t_i{=}1)\wedge(E_i = E_{i-1}),
\end{equation}
removing a further 1.8\% of minutes.

\textbf{Spike detection and smoothing.}
Transient sensor artifact and baseline drift are the dominant nuisance sources in the surviving minutes, and because several of our engineered descriptors are variance-, excursion-, and entropy-based, artifact left in place is read directly as circuit instability. Each channel is therefore partitioned into non-overlapping windows of length $L$, and within each window the local standard deviation is computed. A sample $x_t$ is identified as a spike if it deviates from the window mean by more than $k\sigma$, where $k$ is a threshold and $\sigma$ is the local standard deviation. Rather than fixing $(L,k)$ by visual inspection, which risks tuning the denoiser to the outcome, we select the operating point empirically on a synthetic benchmark with known spike positions (Sec.~\ref{sec:denoise}). Detected spikes are replaced by the mean or median of up to $p$ valid neighboring samples on each side, suppressing transient artifacts while preserving the local physiological trend:
\begin{equation}
x_t \leftarrow \operatorname{mean/median}\{x_j : j \in N_t\},
\end{equation}
where $N_t$ denotes the set of valid neighboring indices surrounding sample $t$.

\subsection{Data Structuring and Derived Physiology}
\label{sec:Structuring}
After the signal is denoised, it undergoes structural formatting to map the continuous stream to the day-wise prediction framework:

\textbf{Gap-aware reconstruction.}
Missing intervals are treated according to what the underlying physiology can support. Short dropouts are linearly interpolated: clot burden and membrane fouling develop over tens of minutes to hours~\cite{sansom2019}, so a brief logging gap cannot span a change in circuit state. Longer gaps and out-of-bounds minutes remain explicitly masked and are propagated to the model as a validity indicator rather than imputed, since interpolating across them would fabricate unobserved physiological trends that a sequence model can exploit. This distinction is what allows the same day to be represented as a fixed-length tensor without silently inventing signal.

\textbf{Derived physiology channels.}
To explicitly capture inter-channel relationships indicative of circuit instability (e.g., membrane fouling and flow resistance), three interpretable pressure gradients are computed at each minute:
\begin{equation}
\mathrm{TMP}=\tfrac{1}{2}(F+R)-E,\quad
\mathrm{PFD}=F-R,\quad
\mathrm{ARD}=R-A,
\end{equation}
where $A$, $F$, $E$, and $R$ denote the access, filter, effluent, and return pressures, respectively. Here, transmembrane pressure (TMP) represents the pressure gradient across the membrane, filter pressure drop (PFD) measures the pressure gradient across the hemofilter, and the access-to-return difference (ARD) captures the pressure difference between the access and return lines~\cite{kdigo2012}. The deep model receives the multivariate signal $X\in\mathbb{R}^{6\times1440}$ over the channels $\{A,F,E,R,\mathrm{TMP},\mathrm{PFD}\}$ together with a validity mask $m$, while $\mathrm{ARD}$ is used exclusively by the engineered-feature branch.

\textbf{Engineered physiology features.}
The engineered branch summarizes each treatment day using 200 standard time-series features (Table~\ref{tab:feats}). A base set (133 features) captures marginal distributions (e.g., central tendency, skewness) and simple within-day dynamics (e.g., linear trends, variability) to describe general pressure levels and coarse volatility~\cite{christ2018}. To explicitly capture the finer circuit-instability dynamics missed by these base statistics, the enriched set (67 features) adds advanced descriptors, including spectral features from the Welch power spectral density~\cite{welch1967}, permutation entropy for nonlinear irregularity~\cite{bandt2002}, excursion statistics, end-of-day deterioration slopes, and cross-channel coupling correlations. All features are computed over valid minutes only, so masked intervals neither contribute to nor dilute the summaries. Prior to modeling, the 200 machine features and the clinical variables are median-imputed, standardized, and reduced by minimum redundancy maximum relevance (mRMR)~\cite{peng2005} selection applied separately within each modality, yielding a grouped budget whose size is treated as a tuned hyperparameter (Sec.~\ref{sec:results}).

\begin{table}[!tb]
\caption{Taxonomy of engineered machine features. Pipelines use the full 200, reduced by mRMR to the selected machine budget (40 in the deployed model).}
\label{tab:feats}
\centering
\scriptsize
\setlength{\tabcolsep}{3.5pt}
\begin{tabular}{p{2.05cm}p{3.05cm}cc c}
\toprule
Family & Features & /ch. & Ch. & N \\
\midrule
\multicolumn{5}{l}{\textit{Base set (133): distribution and within-day dynamics}}\\
Distribution & mean, std, min, max, median, q25/q75, IQR, range, skew, kurtosis, coefficient of variation, valid-minute count & 13 & 7 & 91 \\
Temporal & slope, trend $R^2$, mean $|\Delta|$, max jump, lag-1 autocorr., excursion frac. & 6 & 7 & 42 \\
\cmidrule(l){5-5}
\multicolumn{4}{r}{\textit{Base subtotal}} & 133 \\
\midrule
\multicolumn{5}{l}{\textit{Enriched circuit-instability set (67)}}\\
Spectral & entropy, low/high band power, dominant period~\cite{welch1967} & 4 & 7 & 28 \\
Complexity & permutation entropy~\cite{bandt2002} & 1 & 7 & 7 \\
Excursion & crossing rate, max run above median~\cite{christ2018} & 2 & 7 & 14 \\
End-of-day & delta, slope (last 2\,h) & 2 & 7 & 14 \\
Cross-channel & 3 correlations, co-trend & 4 & -- & 4 \\
\cmidrule(l){5-5}
\multicolumn{4}{r}{\textit{Enriched subtotal}} & 67 \\
\midrule
\multicolumn{4}{r}{\textbf{Total}} & \textbf{200} \\
\bottomrule
\end{tabular}
\end{table}

\begin{figure*}[!t]
\centering
\includegraphics[
    width=\textwidth,
    trim={0.25cm 0.25cm 0.25cm 0.25cm},
    clip
]{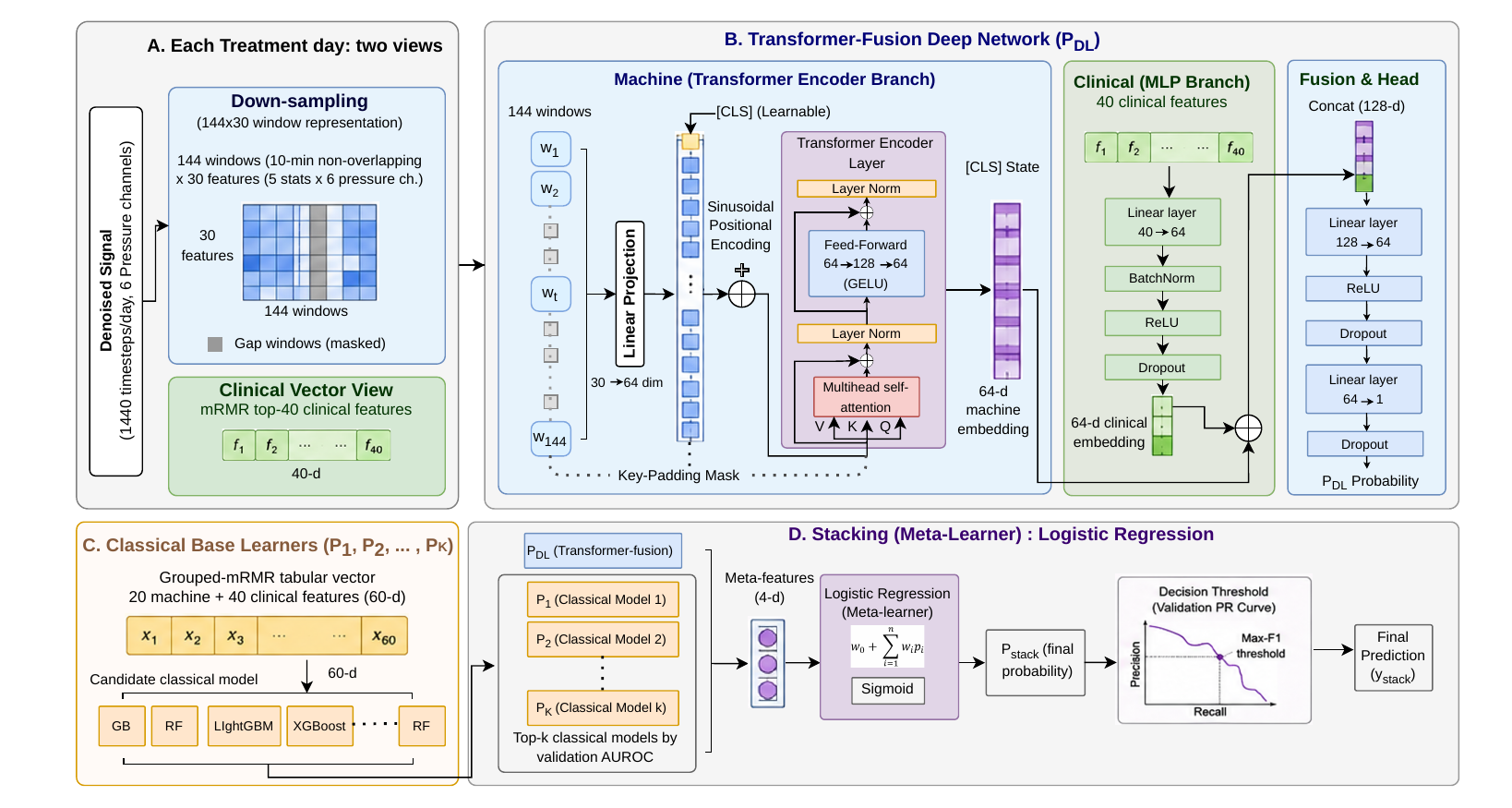}
\caption{Two-level stacked ensemble. (A) Each denoised treatment day yields a $144{\times}30$ window-reduced machine sequence (gap windows masked) and a 60-d mRMR clinical vector. (B) A Transformer encoder with a [CLS] token encodes the sequence, an MLP branch encodes the clinical vector, and a fusion head over the concatenated embeddings yields $P_{\text{DL}}$. (C) Classical learners on the 100-d grouped-mRMR vector (40 machine $+$ 60 clinical) emit $P_1,\dots,P_K$, ranked by validation AUROC. (D) A logistic-regression meta-learner, fit on validation, combines the four probabilities into $P_{\text{stack}}$, thresholded at the validation max-$F_1$ point.}
\label{fig:arch}
\end{figure*}

\subsection{Stacked ensemble of a Transformer-fusion network and classical learners}
\label{sec:proposed}
In the stacked ensemble model(Fig.~\ref{fig:arch}), a logistic-regression meta-learner combines the predicted probabilities from a Transformer-fusion deep network and a pool of classical tabular base learners into the final estimate $P_{\text{stack}}$.

\subsubsection{Input representations (Fig.~\ref{fig:arch}A)}
\label{sec:inputs}
Each 1440-minute denoised circuit-day over the six channels is formatted into three distinct views:

First, the \textbf{machine sequence view} down-samples the stream into 144 non-overlapping 10-minute windows to reduce dimensionality and control overfitting. Computing five statistics (mean, min, max, standard deviation, and least-squares slope) over the valid minutes of each window yields a 30-dimensional token per window:
\begin{equation}
x_{seq} = [x_1, x_2, \ldots, x_{144}] \in \mathbb{R}^{144 \times 30}.
\end{equation}
A window containing no valid minutes is identified by a validity mask ($m_i = 0$) and explicitly masked out of the attention mechanism; otherwise, $m_i = 1$.

Second, the \textbf{clinical vector view} feeds the deep network's clinical branch with the top-60 mRMR-selected patient features (e.g., labs and severity scores), which are standardized and median-imputed:
\begin{equation}
x_{clin} = [f_1, f_2, \ldots, f_{60}] \in \mathbb{R}^{60}.
\end{equation}

Finally, the \textbf{grouped-mRMR tabular view} feeds the classical base learners by concatenating the top-40 engineered machine features with the top-60 clinical features into a single vector:
\begin{equation}
x_{tab} = [\text{machine}_{1..40};\ \text{clinical}_{1..60}] \in \mathbb{R}^{100}.
\end{equation}
The budget $(n_{\mathrm{machine}}, n_{\mathrm{clinical}})$ is selected on validation and swept in Sec.~\ref{sec:results}.


\subsubsection{Transformer-Fusion Deep Network ($P_{\text{DL}}$) (Fig.~\ref{fig:arch}B)}
\label{sec:dl}
The deep network has two branches that are trained jointly. A machine branch encodes the down-sampled window sequence $x_{\text{seq}}$ with a Transformer encoder~\cite{vaswani2017}, and a clinical branch encodes the reduced clinical vector $x_{\text{clin}}$ with a small multilayer perceptron (MLP). The two branch embeddings are concatenated (late fusion) and passed through a fusion head that outputs a single mortality probability $P_{\text{DL}}$. Late fusion keeps each modality separate until it has been encoded on its own, so each branch retains its own inductive bias. The configuration was selected by validation AUROC over $12$ candidate settings: transformer / hidden width $d{=}64$, one encoder layer, $h{=}4$ attention heads, feed-forward width $128$, and dropout $0.4$.

\textbf{Machine branch: Transformer encoder.}
Each $30$-dimensional window token is linearly projected to width $d$, a learnable summary token ($[\text{CLS}]$) is prepended, and sinusoidal positional encodings are added so the encoder is aware of window order. A single encoder layer then lets every valid window attend to every other valid window through multi-head self-attention, followed by a position-wise feed-forward network, each wrapped in a residual connection and layer normalization. Invalid or gap windows are removed from attention by a key-padding mask $M$, so only observed signal contributes; this is the point at which the masking decisions of Sec.~\ref{sec:Structuring} enter the model rather than being smoothed away. The core operation, for head $r$, is scaled dot-product attention over the queries, keys, and values $Q_r,K_r,V_r$ (linear projections of the token sequence, with per-head dimension $d_k=d/h$)~\cite{vaswani2017}:
\begin{equation}
\text{head}_r=\mathrm{softmax}\!\Big(\frac{Q_rK_r^{\!\top}}{\sqrt{d_k}}+M\Big)V_r .
\label{eq:attn}
\end{equation}
The encoded state of the summary token is taken as the $d$-dimensional machine-day embedding $e_{\text{mach}}$.

\textbf{Clinical branch: MLP.}
The clinical vector is mapped to a $d$-dimensional embedding by one fully connected block with batch normalization ($BN$), a ReLU non-linearity, and dropout, which regularizes the low-dimensional clinical input:
\begin{equation}
e_{\text{clin}}=\mathrm{Dropout}\big(\mathrm{ReLU}(\mathrm{BN}(W_c\,x_{\text{clin}}+b_c))\big),
\label{eq:eclin}
\end{equation}
where $W_c,b_c$ are the layer weight and bias.

\textbf{Fusion and head.}
The machine and clinical embeddings are concatenated into a single vector $u=[\,e_{\text{mach}};e_{\text{clin}}\,]$ and passed through a two-layer head (a hidden layer with ReLU and dropout, then a linear output layer) that produces a logit $z_{\text{DL}}$. A logistic sigmoid $\sigma(z)=1/(1+e^{-z})$ turns this into the deep-network probability:
\begin{equation}
P_{\text{DL}}=\sigma(z_{\text{DL}}).
\label{eq:pdl}
\end{equation}

\subsubsection{Classical Base Learners ($P_1, \ldots, P_K$) (Fig.~\ref{fig:arch}C)}
In parallel with the deep network, a pool of tabular models operates on the grouped-mRMR feature vector $x_{tab}$, preprocessed with the same TRAIN-fit imputation and standardization. Five families are considered: Gradient Boosting, Random Forest, LightGBM, XGBoost, and Logistic Regression. Each family is hyperparameter-tuned by randomized search with stratified cross-validation on TRAIN (scored by AUROC) and refit over five seeds, from which the per-day probabilities are combined by taking the maximum across seeds. Each tuned family thus emits a probability:
\begin{equation}
P_k = f_k(x_{tab}).
\end{equation}
The families are ranked by validation AUROC and the top three are carried into the stack ($K=3$ in the reported model).

\subsubsection{Stacking Meta-Learner ($P_{stack}$) (Fig.~\ref{fig:arch}D)}
The deep-network probability (forced in) and the top-three classical probabilities $P_1, \ldots, P_3$ are treated as meta-features and combined by a logistic-regression meta-learner (stacked generalization). Writing the meta-feature vector as $\phi = [\, P_{DL},\, P_1,\, P_2,\, P_3 \,]$, the final stacked probability is:
\begin{equation}
P_{stack} = \sigma\!\left(w^\top \phi + b\right),
\end{equation}
where $w$ and $b$ are the learnable meta-learner weight vector and bias, fit on the validation probabilities and labels rather than the test set, so the stacking weights cannot leak test information. For binary decisions, the operating threshold is selected on the validation set by maximizing $F_1$ score.

\subsection{Training, Selection, and Leakage Control}
The deep network is trained with a class-weighted binary cross-entropy loss on its output probability $p=\sigma(z)$:
\begin{equation}
\mathcal{L}=-\frac{1}{N}\sum_{i=1}^{N}\big[\,w_{+}\,y_i\log p_i+(1-y_i)\log(1-p_i)\,\big],
\label{eq:bce}
\end{equation}
where $w_{+}$ upweights the minority death class. Imbalance is additionally addressed at the sampling level by a weighted random sampler that oversamples deaths into approximately balanced minibatches, countering the $\sim$1:8 day-level class ratio. Optimization uses AdamW (initial learning rate $3\!\times\!10^{-4}$, decayed toward $10^{-6}$; weight decay $5\!\times\!10^{-3}$) with dropout $0.4$, gradient-norm clipping at $5.0$, and batch size $64$, for up to $60$ epochs with early stopping on validation AUROC and a best-of-five-seeds checkpoint.

A strict fit-on-split barrier controls leakage. The train split fits all imputation, standardization, feature selection, neural weights, and classical-model tuning. The validation split fits the base-learner ranking, the stacking meta-learner, the decision threshold, and the early-stopping and best-seed choices. The test split is untouched during fitting and is used only to report final performance.

\section{Experiments \& Results}
Classical models underwent a 30-iteration randomized search with 5-fold cross-validation. Deep architectures were trained over five seeds. The primary evaluation metric is the day-level test AUROC. Secondary metrics (AUPRC, F1, precision, and recall)~\cite{boyd2013} were computed using the threshold that maximized the F1 score on the validation split. Performance was scored under full, 1:1, and 1:2 prevalence protocols.

\subsection{Denoising validation}
\label{sec:denoise}
\begin{figure}[!t]
\centering
\includegraphics[
    width=0.85\columnwidth,
    trim={8.75cm 2.30cm 11.35cm 0.55cm},
    clip
]{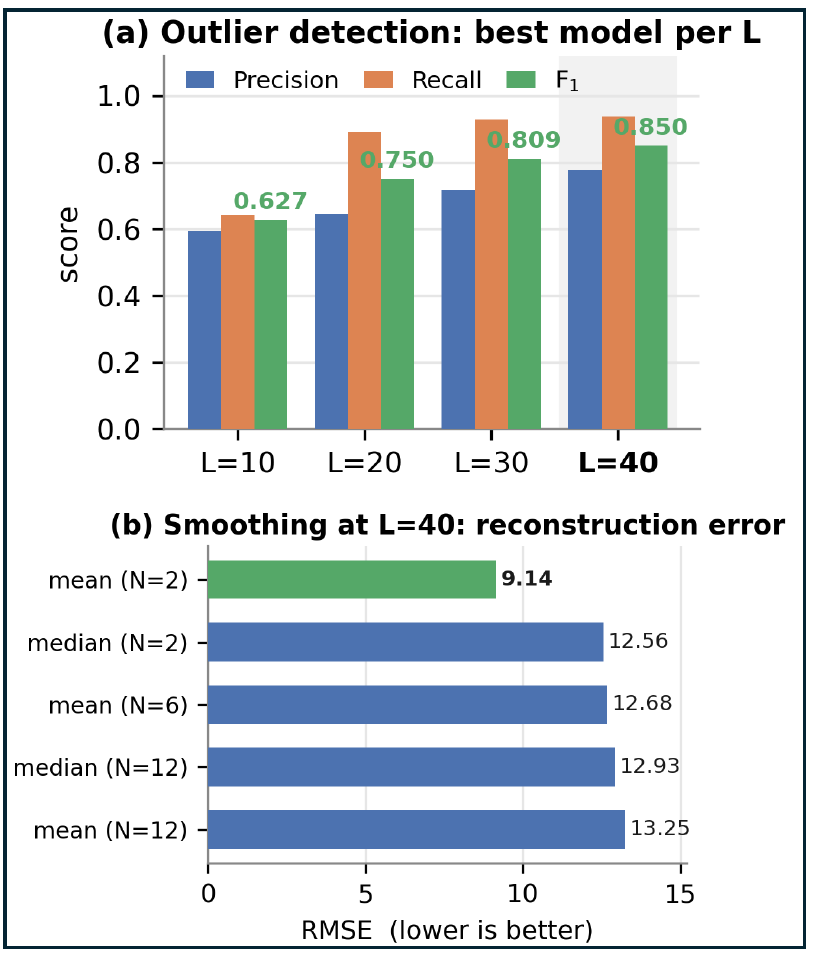}
\caption{Peak-detection performance on CRRT pressure waveforms.
(a)~Precision, recall, and $F_1$ for the best outlier-detection model at each
window length $L$ ($F_1$ peaks at $L=40$, $F_1=0.850$). (b)~Reconstruction
error for smoothing variants at $L=40$ (mean $N=2$ is lowest, RMSE $9.14$).}
\label{fig:peak_detection}
\end{figure}

The peak-denoising operating point was fixed on a synthetic spike-injection benchmark (Fig.~\ref{fig:flow}B), so that the denoiser is tuned against known ground truth rather than against the downstream outcome. Fifteen 180-minute raw chunks were manually cleared of spikes to serve as ground truth, then corrupted by injecting spikes into 5\% of samples at known positions. A grid search was run over the detection parameters (window size $L \in \{20,30,40,50,60\}$, standard-score threshold $\kappa \in \{2,3\}$) and the smoothing parameters (number of valid neighbors $\eta \in \{2,4,6,8,10\}$, aggregation by mean or median). Each configuration was scored on spike-detection quality against the known spike positions (precision, recall, $F_1$) and on recovery error against the clean signal (MAE, MSE, RMSE)~\cite{chai2014}. As shown in Fig.~\ref{fig:peak_detection}, detection quality improves monotonically with window length over the range searched, and the best setting was $L=40$, $\kappa=2$, with mean smoothing over $\eta=2$ neighbors, which maximized detection $F_1$ ($0.850$) while minimizing reconstruction error (RMSE $9.14$). This operating point is then held fixed for every model and every pipeline reported below.

\subsection{Three-Pipeline Benchmark}
\label{sec:results}
We evaluate the Windowed-Transformer Fusion Stack against established deep sequence-modeling baselines spanning convolutional, recurrent, and attention-based families (Table~\ref{tab:dlbench}) and eight classical learners (Table~\ref{tab:classical}), across the clinical-only, machine-only, and combined pipelines. The convolutional, recurrent, and attention-pooling encoders consume the $144{\times}30$ window-reduced representation of Sec.~\ref{sec:inputs} with its validity mask, whereas the Patch-Transformer baseline consumes the full 1440-minute stream in non-overlapping 60-minute patches; the contrast with our deployed encoder, which applies the same backbone to the window-reduced tokens, therefore isolates the input representation from the architecture. Each encoder is late-fused with an identical clinical MLP branch, and the clinical MLP alone serves as the EHR-only reference. All neural models train under the shared protocol of Sec.~\ref{sec:proposed}, so differences across pipelines reflect the input stream rather than the training regime.

\begin{figure}[!t]
\centering
\includegraphics[
    width=\columnwidth,
    trim={3.6cm 2.5cm 3.4cm 2.5cm},
    clip
]{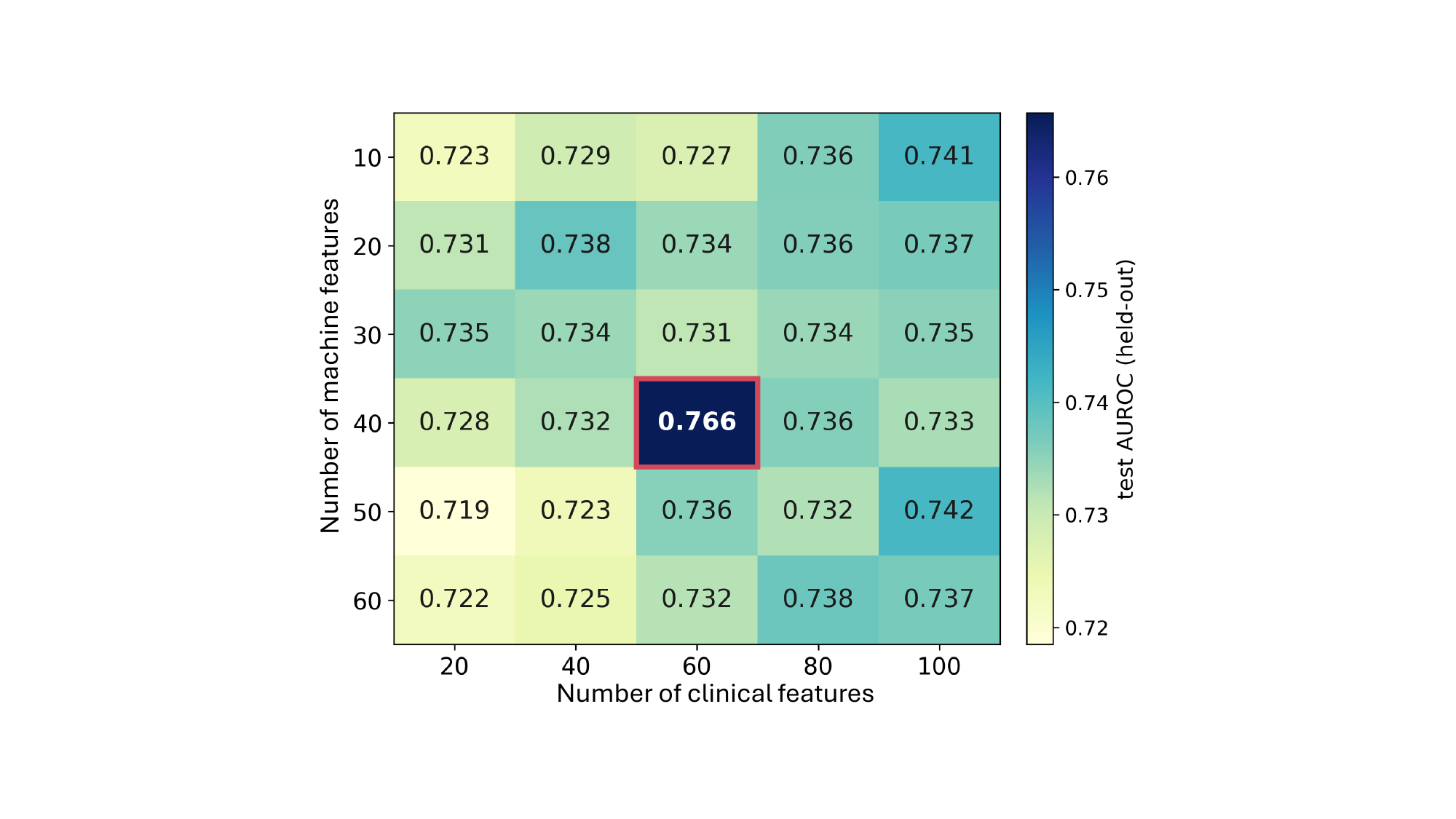}
\caption{Held-out test AUROC of the Windowed-Transformer Fusion Stack
across the grouped-mRMR feature budget
($n_{\mathrm{machine}} \times n_{\mathrm{clinical}}$).
The red box marks the validation-selected optimum,
 $40$ machine $\times$ $60$ clinical.}
\label{fig:heatmap}
\end{figure}

Because the grouped-mRMR budget is itself a design choice that could confound any pipeline comparison, we first swept it over $n_{\mathrm{machine}} \in \{10,\dots,60\}$ and $n_{\mathrm{clinical}} \in \{20,\dots,100\}$ (Fig.~\ref{fig:heatmap}). The budget was selected on validation; Fig.~\ref{fig:heatmap} reports test AUROC only as a sensitivity check. Values vary within a narrow band ($0.719$–$0.766$) with no trend in either axis, so the reported result is not an artifact of feature count. All classical and deep models in Tables~\ref{tab:dlbench} and~\ref{tab:classical} are evaluated at this common budget, so no comparison below is confounded by feature count.

The clinical-only pipeline sets the reference an EHR-driven system can reach, led by Gradient Boosting (AUROC $0.717$, Table~\ref{tab:classical}) and the clinical MLP ($0.710$). The machine-only pipeline addresses our first question: with no clinical data, the waveform alone yields an AUROC of $0.625$. Our ensemble exceeds every classical learner on the same features (best $0.608$, Table~\ref{tab:classical}) and every standalone deep encoder ($0.579$--$0.584$). This substantiates the core assumption of the study, that minute-level circuit dynamics carry patient-level mortality information independent of the EHR.

The combined pipeline addresses our second question. Fusing both streams reaches an AUROC of $0.766$, a lift of $+0.049$ over the best clinical-only model, $+0.038$ over the best combined classical learner ($0.728$), and $+0.084$ over the best deep fusion baseline (Attention-CNN1D, $0.682$). The waveform is thus not redundant with routine labs, and the stacked design is what converts that complementarity into a measurable gain. The gain is in discrimination, not precision: under full imbalance the stack's AUPRC (0.216) is comparable to the best clinical-only learners (0.233, 0.222; Table~\ref{tab:classical}) against a 0.070 floor, a difference not resolvable on 62 positives. Once the positive class is better populated the ordering separates, with the stack highest in AUPRC under both $1$:$1$ ($0.756$) and $1$:$2$ ($0.616$).

\begin{figure}[!t]
\centering
\includegraphics[
    width=0.82\columnwidth,
    trim={5.95cm 3.55cm 8.30cm 2.80cm},
    clip
]{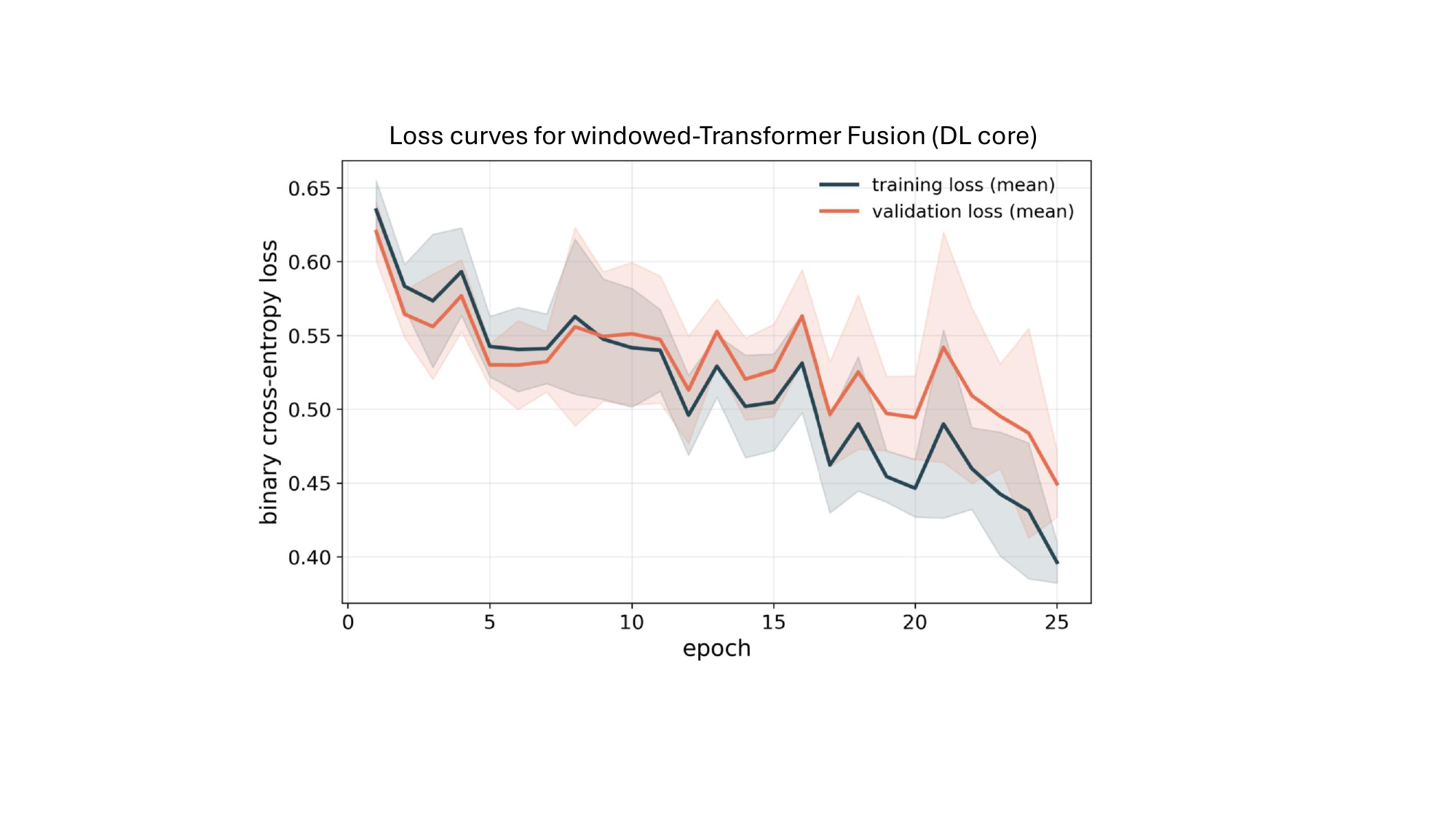}
\caption{Training and validation loss curves for the deployed combined fusion model (one-day mortality, class-weighted BCE, mean over five seeds with shaded range). The small train--validation gap indicates effective regularization; curves are truncated at the early-stopping patience limit.}
\label{fig:loss_curve}
\end{figure}

\begin{table*}[!tb]
\caption{Deep-learning benchmark across clinical only, machine only and combined pipeline.}
\label{tab:dlbench}
\centering
\scriptsize
\setlength{\tabcolsep}{4pt}
\begin{tabular}{lccccccccc}
\toprule
& \multicolumn{5}{c}{Full (imbalanced) test} & \multicolumn{2}{c}{Balanced 1:1} & \multicolumn{2}{c}{Balanced 1:2}\\
\cmidrule(lr){2-6}\cmidrule(lr){7-8}\cmidrule(lr){9-10}
Model & AUROC & AUPRC & F1 & Prec. & Rec. & AUROC & AUPRC & AUROC & AUPRC\\
\midrule
\multicolumn{10}{l}{\textit{Clinical-only}}\\
Clinical MLP & 0.710 & 0.177 & 0.242 & 0.229 & 0.258 & 0.717 & 0.716 & 0.712 & 0.561 \\
\midrule
\multicolumn{10}{l}{\textit{Machine-only}}\\
CNN1D\textsuperscript{a}~\cite{wang2017tsc} & 0.579 & 0.095 & 0.026 & 0.071 & 0.016 & 0.575 & 0.585 & 0.579 & 0.415 \\
RNN (BiLSTM/GRU)\textsuperscript{a}~\cite{cho2014gru}  & \underline{0.584} & 0.094 & 0.159 & 0.098 & 0.419 & \underline{0.583} & 0.583 & \underline{0.585} & 0.412 \\
Patch-Transformer\textsuperscript{b}~\cite{nie2023patchtst} & 0.581 & 0.086 & 0.123 & 0.099 & 0.161 & 0.578 & 0.559 & 0.582 & 0.389 \\
Attention-CNN1D\textsuperscript{a} & 0.579 & \underline{0.102} & 0.165 & 0.155 & 0.177 & 0.578 & \underline{0.601} & 0.579 & \underline{0.430} \\
Windowed-Transformer Ensemble \textit{(ours)}\textsuperscript{a,c} & \textbf{0.625} & \textbf{0.109} & 0.159 & 0.109 & 0.290 & \textbf{0.624} & \textbf{0.611} & \textbf{0.628} & \textbf{0.445} \\
\midrule
\multicolumn{10}{l}{\textit{Combined}}\\
CNN1D fusion \textsuperscript{a}~\cite{wang2017tsc} & 0.646 & 0.182 & 0.188 & 0.121 & 0.419 & 0.644 & 0.674 & 0.646 & 0.526 \\
RNN (BiLSTM/GRU) fusion \textsuperscript{a}~\cite{cho2014gru} & 0.669 & \textbf{0.222} & 0.278 & 0.302 & 0.258 & 0.667 & \underline{0.698} & 0.669 & \underline{0.559} \\
Patch-Transformer fusion \textsuperscript{b}~\cite{nie2023patchtst} & 0.680 & 0.182 & 0.225 & 0.156 & 0.403 & \underline{0.680} & 0.696 & \underline{0.682} & 0.549 \\
Attention-CNN1D fusion\textsuperscript{a} & \underline{0.682} & 0.156 & 0.239 & 0.163 & 0.452 & \underline{0.680} & 0.692 & 0.681 & 0.536 \\
Windowed-Transformer Fusion Stack \textit{(ours)}\textsuperscript{a,d} & \textbf{0.766} & \underline{0.216} & 0.296 & 0.263 & 0.339 & \textbf{0.765} & \textbf{0.756} & \textbf{0.766} & \textbf{0.616} \\
\bottomrule
\end{tabular}

\vspace{1pt}
\parbox{\textwidth}{\footnotesize \textit{Note:} \textsuperscript{a}$144{\times}30$ window-reduced sequence with validity mask. \textsuperscript{b}Full 1440-min stream in 60-min patches. \textsuperscript{c}Top-three classical models $+$ windowed Transformer (machine mRMR-40). \textsuperscript{d}Windowed-Transformer fusion $+$ top-three classical models, logistic-stacked (40 machine $+$ 60 clinical grouped-mRMR). All rows use the same validation-selected feature budget (Fig.~\ref{fig:heatmap}). AUROC is primary because it is prevalence-invariant and thus comparable across the three protocols; with 62 positive days, AUPRC differences below $\sim$0.03 are not resolvable; F1/precision/recall use one validation-selected threshold and collapse under imbalance. Best AUROC/AUPRC per block in \textbf{bold}, second \underline{underlined}.}
\end{table*}

\begin{table*}[!tb]
\caption{Classical benchmark across all three pipelines (one-day horizon; per-day, current denoised).
Full imbalanced test with prevalence-balanced 1:1 and 1:2 resamples (mean of 100).
Threshold $=$ max-$F_1$ on validation. Feature budget: clinical-only $=$ 60 clinical, machine-only $=$
40 machine, combined $=$ 40 machine $+$ 60 clinical (grouped-mRMR).}
\label{tab:classical}
\centering
\scriptsize
\setlength{\tabcolsep}{4pt}
\begin{tabular}{lccccccccc}
\toprule
& \multicolumn{5}{c}{Full (imbalanced) test} & \multicolumn{2}{c}{Balanced 1:1} & \multicolumn{2}{c}{Balanced 1:2}\\
\cmidrule(lr){2-6}\cmidrule(lr){7-8}\cmidrule(lr){9-10}
Model & AUROC & AUPRC & F1 & Prec. & Rec. & AUROC & AUPRC & AUROC & AUPRC\\
\midrule
\multicolumn{10}{l}{\textit{Clinical-only}}\\
LogReg       & 0.710 & 0.222 & 0.264 & 0.232 & 0.306 & 0.714 & 0.726 & 0.710 & 0.583 \\
DecisionTree & 0.617 & 0.117 & 0.215 & 0.177 & 0.274 & 0.614 & 0.608 & 0.614 & 0.444 \\
RandomForest & 0.711 & \textbf{0.233} & 0.239 & 0.173 & 0.387 & 0.713 & \textbf{0.731} & 0.709 & \textbf{0.589} \\
GradBoost    & \textbf{0.717} & 0.209 & 0.277 & 0.203 & 0.435 & \textbf{0.720} & 0.721 & \textbf{0.716} & 0.575 \\
AdaBoost     & 0.702 & 0.186 & 0.284 & 0.264 & 0.306 & 0.704 & 0.713 & 0.700 & 0.562 \\
KNN          & 0.590 & 0.151 & 0.159 & 0.156 & 0.161 & 0.595 & 0.625 & 0.589 & 0.468 \\
LightGBM     & \underline{0.716} & 0.209 & 0.265 & 0.190 & 0.435 & \underline{0.719} & \underline{0.730} & \underline{0.715} & \underline{0.584} \\
XGBoost      & 0.685 & \underline{0.217} & 0.246 & 0.235 & 0.258 & 0.687 & 0.710 & 0.684 & 0.569 \\
\midrule
\multicolumn{10}{l}{\textit{Machine-only}}\\
LogReg       & 0.589 & \underline{0.109} & 0.102 & 0.139 & 0.081 & 0.589 & \textbf{0.610} & 0.587 & \textbf{0.439} \\
DecisionTree & 0.552 & 0.085 & 0.131 & 0.070 & 1.000 & 0.551 & 0.542 & 0.553 & 0.377 \\
RandomForest & 0.584 & \underline{0.109} & 0.098 & 0.078 & 0.129 & 0.582 & 0.587 & 0.581 & 0.418 \\
GradBoost    & \underline{0.598} & 0.101 & 0.153 & 0.097 & 0.371 & \underline{0.598} & \underline{0.601} & \underline{0.598} & 0.429 \\
AdaBoost     & \textbf{0.608} & 0.095 & 0.115 & 0.117 & 0.113 & \textbf{0.605} & 0.585 & \textbf{0.609} & 0.416 \\
KNN          & 0.559 & 0.089 & 0.159 & 0.099 & 0.403 & 0.561 & 0.565 & 0.560 & 0.394 \\
LightGBM     & 0.603 & 0.107 & 0.100 & 0.132 & 0.081 & 0.604 & 0.604 & 0.602 & \underline{0.435} \\
XGBoost      & 0.587 & 0.105 & 0.147 & 0.084 & 0.581 & 0.587 & \underline{0.601} & 0.586 & \underline{0.435} \\
\midrule
\multicolumn{10}{l}{\textit{Combined}}\\
LogReg       & \underline{0.725} & \textbf{0.218} & 0.254 & 0.198 & 0.355 & \underline{0.726} & \textbf{0.736} & \underline{0.724} & \textbf{0.593} \\
DecisionTree & 0.535 & 0.096 & 0.176 & 0.139 & 0.242 & 0.534 & 0.575 & 0.534 & 0.408 \\
RandomForest & \textbf{0.728} & 0.188 & 0.240 & 0.166 & 0.435 & \textbf{0.726} & 0.718 & \textbf{0.727} & 0.567 \\
GradBoost    & 0.718 & 0.173 & 0.277 & 0.253 & 0.306 & 0.718 & 0.711 & 0.718 & 0.558 \\
AdaBoost     & 0.698 & 0.159 & 0.242 & 0.168 & 0.435 & 0.700 & 0.700 & 0.699 & 0.543 \\
KNN          & 0.568 & 0.102 & 0.164 & 0.143 & 0.194 & 0.570 & 0.581 & 0.568 & 0.418 \\
LightGBM     & 0.694 & 0.179 & 0.217 & 0.151 & 0.387 & 0.698 & 0.705 & 0.696 & 0.556 \\
XGBoost      & 0.720 & \underline{0.182} & 0.252 & 0.168 & 0.500 & 0.720 & \underline{0.725} & 0.720 & \underline{0.580} \\
\bottomrule
\end{tabular}
\end{table*}

\subsection{Ablation Study}
We ablated the deployed combined model along two axes: the preprocessing applied to the pressure signal and the sequence encoder inside the fusion network (Table~\ref{tab:ablation}).

\textbf{Preprocessing.} Cumulatively applying the cleaning modules raised the
AUROC from $0.737$ on the raw signal to $0.766$ ($+0.029$), with peak removal
the largest single step ($+0.019$). Each individual increment is smaller than
the $\pm0.06$ test confidence interval (62 positive days), so we do not credit
any one module in isolation. The supportable claim is cumulative: the full
pipeline improves AUROC by about as much as the encoder choice below, making
signal conditioning integral to the method rather than a preliminary. The enriched
feature set is the one non-monotonic step ($-0.007$). Its contribution is to
attribution rather than discrimination: it raises the machine share of the top-15
SHAP features from 20.0\% to 46.7\% (Sec.~\ref{sec:interp}) at no measurable cost
in AUROC.

\textbf{Sequence encoder.} With preprocessing fixed to the fully cleaned signal, the windowed Transformer was the strongest encoder ($0.766$), ahead of the Patch-Transformer ($0.732$), CNN1D ($0.729$), and RNN ($0.713$). Because the Patch-Transformer applies the same backbone to 60-minute patches of the raw stream, the $0.034$ gap between the two variants attributes the gain to the window-reduced representation rather than to self-attention alone.

\begin{table}[t]
\centering
\caption{Ablation of the deployed combined model (1-day; 40 machine $\times$ 60 clinical
grouped-mRMR features), along two axes: signal preprocessing and sequence encoder.
Held-out test AUROC; best in \textbf{bold}.}
\label{tab:ablation}
\footnotesize
\setlength{\tabcolsep}{4pt}
\begin{tabular}{l c}
\toprule
Configuration & AUROC \\
\midrule
\multicolumn{2}{l}{\emph{Cumulative configuration ladder (windowed-Transformer encoder)}} \\
Raw signal                                   & 0.737 \\
\quad +  Enriched features (feature set)                    & 0.730 \\
\quad + Priming \& downtime removal          & 0.747 \\
\quad + Peak removal (full preprocessing)    & \textbf{0.766} \\
\midrule
\multicolumn{2}{l}{\emph{Sequence encoder (full preprocessing)}} \\
CNN1D                                         & 0.729 \\
RNN                                           & 0.713 \\
Patch-Transformer                            & 0.732 \\
Windowed Transformer (ours)                  & \textbf{0.766} \\
\bottomrule
\end{tabular}
\end{table}

\subsection{Training Dynamics}
\label{sec:training}
All deep models were trained for up to 60 epochs with AdamW under a
class-weighted binary cross-entropy objective and a weighted sampler
countering the $\sim$1:8 class imbalance, with the learning rate decayed from
$3\times10^{-4}$ toward $10^{-6}$ and early stopping on validation AUROC.
Models were trained on a lab workstation equipped with two NVIDIA RTX 6000 Ada
GPUs (48\,GB VRAM each) and 512\,GB of system RAM. Fig.~\ref{fig:loss_curve}
plots the mean training and validation loss across the five seeds for the
deployed combined fusion core, with shaded bands showing the across-seed
spread. Both losses decrease throughout training, with validation tracking
training closely (gap $\approx0.03$--$0.05$) and widening only modestly after
epoch 17, indicating the fusion core is well regularized; the best-validation
checkpoint is selected per seed by early stopping.


\subsection{Interpretability and Clinical Significance}
\label{sec:interp}
\begin{figure}[!tb]
\centering
\includegraphics[
  width=\columnwidth,
  trim={6.7cm 0.0cm 6.7cm 0.0cm},
  clip
]{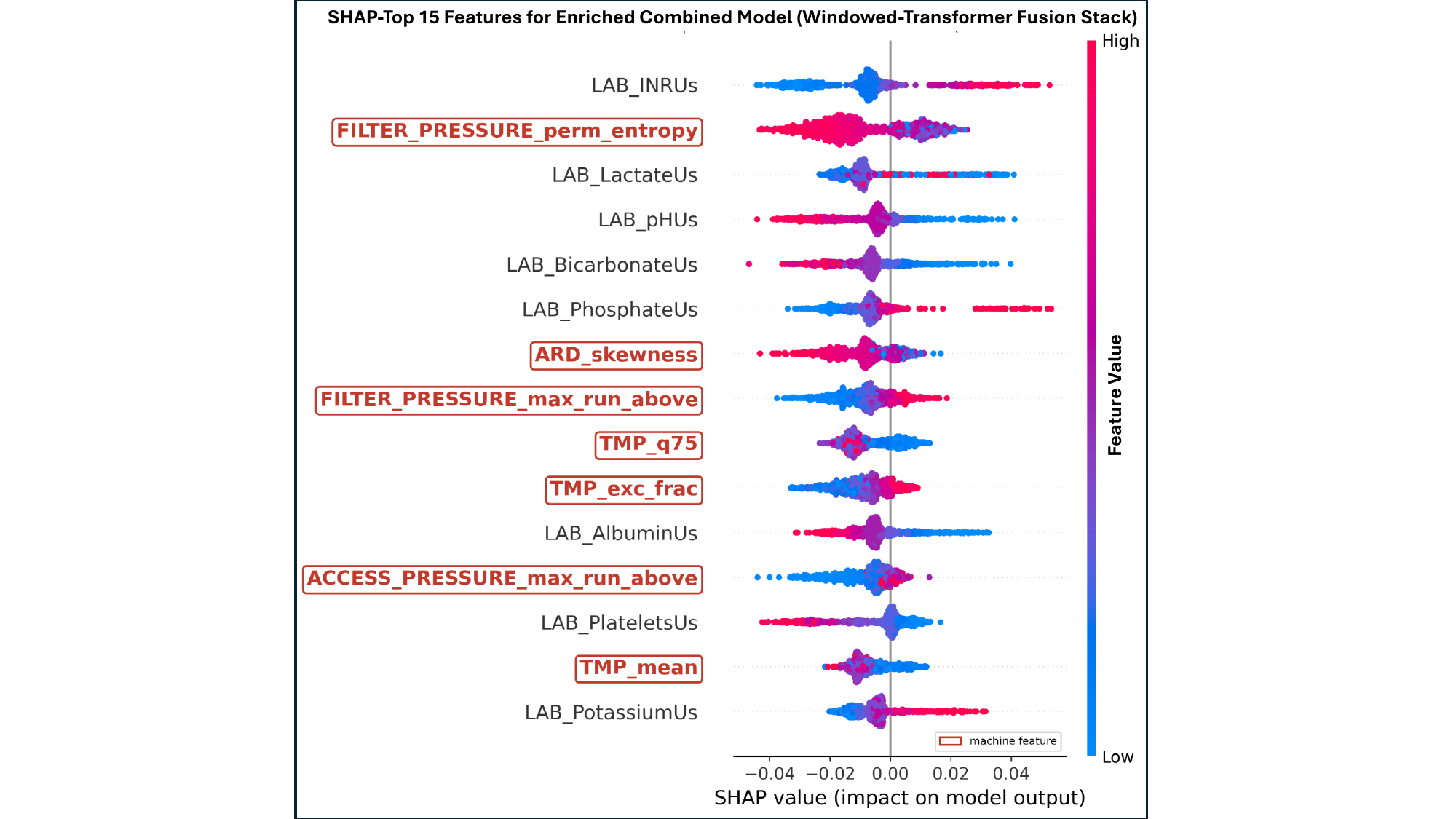}
\caption{SHAP beeswarm (top-15 by mean $|$SHAP$|$) for the Windowed-Transformer Fusion Stack, attributed via its Random-Forest tabular backbone. Red-boxed labels mark CRRT machine descriptors; positive values increase predicted mortality risk.}
\label{fig:shap}
\end{figure}

To assess whether the machine signal is physiologically coherent rather than
an incidental correlate, we computed SHAP attributions on the held-out test
days. Since the stacked probability combines base-learner outputs rather than input
features, attributions are computed on the stack's Random-Forest tabular learner,
which consumes the same grouped-mRMR vector. Adding the enriched circuit-instability descriptors shifts the model's predictive focus toward the waveform: machine features rise from 3 to 7 of the top 15 by mean $|$SHAP$|$ ($20.0\%$ to $46.7\%$), and from rank 5 to rank 2 at best (Table~\ref{tab:shapmachine}). The beeswarm plot (Fig.~\ref{fig:shap}) shows these features interleaved with standard markers of systemic
decompensation rather than clustered below them.

The clinical features that dominate the ranking have clinical relevance in mortality prediction. Elevated INR and low platelets mark severe coagulopathy ~\cite{taylor2001}; elevated lactate, low bicarbonate and low pH mark tissue hypoperfusion and acid--basederangement ~\cite{evans2021}; low albumin, and phosphate derangements complete a picture of inflamation and multi-organ decompensation ~\cite{soeters2019}. That the model recovers this established risk profile without being told it is a validity check: the learned ranking reproduces bedside reasoning before it adds anything to it.

What it adds is the machine signal, and the specific descriptors that surface
carry mechanistic meaning. Filter-pressure permutation entropy, a nonlinear
measure of signal irregularity, ranks second overall, ahead of lactate and pH;
filter-pressure and access-pressure excursion runs, TMP elevation
(\textsc{tmp\_q75}, \textsc{tmp\_mean}), and ARD skewness also enter the top
15 (Fig.~\ref{fig:shap}). Each maps onto a recognized failure process:
erratic filter pressure and sustained TMP elevation are signatures of
progressive micro-clotting and membrane fouling~\cite{sansom2019}, and
access-pressure instability reflects vascular-access dysfunction. In the
beeswarm, high values of these descriptors push predictions toward death,
which is the direction bedside physiology predicts. 

The clinical significance follows from sampling: the dominant labs are drawn
once or twice daily, while the circuit descriptors update every minute, so
the waveform is the only input that can register deterioration between blood
draws. Our results are consistent with this: the waveform alone is weakly
prognostic (AUROC $0.625$, Table~\ref{tab:dlbench}), every fusion model beats
its clinical-only counterpart (Sec.~\ref{sec:results}). At the
bedside, the waveform does not replace laboratory risk assessment but
interpolates it, continuously tracking the same deterioration the next blood
draw would confirm.

\begin{table}[!tb]
\caption{Machine features in the combined-model SHAP ranking: counts among the
top-5/10/15 by mean $|\text{SHAP}|$, and the machine share of the top-15.
Enriched $=$ deployed optimal model (40 machine $+$ 60 clinical grouped-mRMR).}
\label{tab:shapmachine}
\centering
\footnotesize
\setlength{\tabcolsep}{4pt}
\begin{tabular}{lccccc}
\toprule
Combined model & Best rank & top-5 & top-10 & top-15 & \% top-15 \\
\midrule
Base     & 5 & 1 & 2 & 3 & 20.0 \\
Enriched & 2 & 1 & 5 & 7 & 46.7 \\
\bottomrule
\end{tabular}
\end{table}

\section{Discussion and Conclusion}

This study shows that minute-level CRRT circuit pressure streams carry
standalone prognostic information for short-horizon mortality (machine-only
AUROC $0.625$), from a signal that is currently discarded. Fusing the waveform
with routine EHR features raises performance to $0.766$, above the best
clinical-only model ($0.717$), and SHAP attribution locates the gain in
interpretable circuit dynamics with seven CRRT metrics among the top 15 features . Equally important for reuse, the ablation of Table~\ref{tab:ablation} indicates that this result is not obtainable from the raw stream: conditioning the signal is worth $+0.029$ AUROC, on the same order as the encoder choice, so the acquisition and cleaning protocol is part of the finding rather than a preliminary to it. These results support the premise that circuit-instability physiology is a usable dynamic input for rolling risk estimation, not that it is sufficient for it.

Several limitations point toward the next steps. External validation on an
independent cohort is required before any claim of clinical utility, since
CRRTnet is multicenter but our split is stratified at the patient rather than
the site level. To this end, our next step is validation on a single-center
cohort of AKI patients on CRRT at the University of Alabama at Birmingham
(2012--2025), for which both linked EHR and CRRT machine data are available;
because this cohort was collected independently of CRRTnet, it provides a
genuine test of transportability for both the signal pipeline and the model.
The label structure imposes two further constraints. Day-level mortality is
rare ($\sim$6.4\% of treatment days, 62 positives in the test split), which
widens confidence intervals and limits how finely competing configurations can
be distinguished. More fundamentally, the label is coarse: an entire
1440-minute day shares a single outcome, although the physiology that drives
that outcome is unlikely to be spread uniformly across it. The informative
intervals within a day are unknown, and annotating them manually at scale is
not feasible at bedside. Beyond aggregate discrimination, a
rolling bedside score also needs to signal when it can be trusted: the
AKI-CRRT population is heterogeneous, and a model that performs adequately
overall may be unreliable within specific patient phenotypes. Establishing
per-phenotype reliability, so that predictions are surfaced only where the
model is competent, is a necessary step toward the trustworthiness that
decision support demands.

Given its bounded performance, the system is not yet ready for standalone
decision support, but our results show the potential of CRRT data to improve point-of-care decision support. Encoders that consume the full day plateau at $0.579$--$0.584$ on the machine stream alone, while the windowed representation outperforms full-sequence patching. Both point to the same explanation: the deterioration signal is concentrated in short intervals, and averaging over a mostly stable day dilutes it. Since these intervals cannot be hand-annotated, the next step is a weakly supervised model that learns from the day-level label alone which intervals matter. This would recover the diluted signal and, more importantly, indicate when within a day risk rises, a far more actionable output at the bedside than a single daily score. Establishing that the waveform is informative, as this work does, is the precondition for that next step.

\vspace{2pt}
\noindent\textbf{Acknowledgment.} JAN is supported by grants from NIH-NIDDK (R01DK128208, R01DK133539,
U01DK12998, and U54DK137307).

\noindent\textbf{Author Contributions.} JAN and JC conceived and designed the study, supervised the work, and provided clinical and methodological guidance. SIP developed the signal processing pipeline, conducted the model development and experiments, performed all the analysis, and drafted the manuscript. SLG designed and led the CRRTnet study and provided the data. JY, JL, LC, TKC, and GNN contributed to critical revision of the manuscript. All authors reviewed and approved the final manuscript.

\noindent\textbf{Funding.} This work was supported by NIH-NIDDK R01DK133539 (MPI GNN and JAN). The funding federal agency was not involved in the design or conduction of the study, and this report does not represent official views of the NIH. The CRRTnet study was funded by an investigator-initiated grant from Vantive US Healthcare, LLC (PI: SLG). Neither Vantive nor its personnel had any involvement in the design of CRRTnet or the collection, analysis, and interpretation of data; writing the report; and the decision to submit the report for publication.

\noindent\textbf{Ethics Approval and Consent to Participate.} The CRRTnet study was approved by the local institutional review board at
each participating site, with a waiver of informed consent granted due to
the deidentified and observational nature of the data.

\begingroup
\footnotesize
\emergencystretch=1em
\setlength{\itemsep}{0pt plus 0.3ex}

\endgroup


\begin{thebibliography}{00}

\bibitem{russo2019} D. S. Russo \textit{et al.}, ``Comparison of hemodynamic parameters among continuous, intermittent and hybrid renal replacement therapy in acute kidney injury: a systematic review of randomized clinical trials,'' 2019.

\bibitem{negi2016} S. Negi, D. Koreeda, S. Kobayashi, Y. Iwashita, and T. Shigematsu, ``Renal replacement therapy for acute kidney injury,'' \textit{Renal Replacement Therapy}, vol. 2, pp. 1--7, 2016.

\bibitem{negi2023} S. Negi, T. Wada, N. Matsumoto, J. Muratsu, and T. Shigematsu, ``Current therapeutic strategies for acute kidney injury,'' \textit{Renal Replacement Therapy}, vol. 9, pp. 1--7, 2023.

\bibitem{patel2022} A. K. Patel, E. Trujillo-Rivera, H. Morizono, and M. Pollack, ``The Criticality Index-Mortality: A dynamic machine learning prediction algorithm for mortality prediction in children cared for in an ICU,'' \textit{Front. Pediatr.}, vol. 10, 2022.

\bibitem{notaro2026} S. Notaro \textit{et al.}, ``Predicting failure of extubation and non-invasive respiratory support in critically ill patients: clinical complexity, limitations of traditional indices, and machine learning perspectives,'' \textit{Front. Med.}, vol. 13, 2026.

\bibitem{deasy2020} J. Deasy, P. Li\`{o}, and A. Ercole, ``Dynamic survival prediction in intensive care units from heterogeneous time series without the need for variable selection or curation,'' \textit{Sci. Rep.}, vol. 10, no. 1, p. 22129, 2020.

\bibitem{choi2022} M. H. Choi \textit{et al.}, ``Mortality prediction of patients in intensive care units using machine learning algorithms based on electronic health records,'' \textit{Sci. Rep.}, vol. 12, no. 1, p. 7180, 2022.

\bibitem{sansom2019} B. Sansom, S. Sriram, J. Presneill, and R. Bellomo, ``Circuit hemodynamics and circuit failure during continuous renal replacement therapy,'' \textit{Crit. Care Med.}, vol. 47, no. 11, pp. e872--e879, 2019.

\bibitem{yang2024} E. Yang \textit{et al.}, ``Development and external validation of a prediction model for premature circuit clotting of CRRT,'' \textit{Intensive Crit. Care Nurs.}, vol. 84, p. 103703, 2024.

\bibitem{kang2020} M. W. Kang \textit{et al.}, ``Machine learning algorithm to predict mortality in patients undergoing continuous renal replacement therapy,'' \textit{Crit. Care}, vol. 24, no. 1, p. 42, 2020.

\bibitem{gu2024} M. Gu \textit{et al.}, ``Using machine learning to predict the risk of short-term and long-term death in AKI patients after commencing CRRT,'' \textit{BMC Nephrol.}, vol. 25, no. 1, p. 245, 2024.

\bibitem{thadani2024} S. Thadani \textit{et al.}, ``Machine learning-based prediction model for ICU mortality after CRRT initiation in children,'' \textit{Crit. Care Explor.}, vol. 6, no. 12, e1188, 2024.

\bibitem{zhong2024} L. Zhong \textit{et al.}, ``Risk prediction models for successful discontinuation in AKI undergoing CRRT,'' \textit{iScience}, vol. 27, no. 8, p. 110397, 2024.

\bibitem{zhu2024} S. Zhu \textit{et al.}, ``Machine learning-aided decision-making model for the discontinuation of continuous renal replacement therapy,'' \textit{Blood Purif.}, vol. 53, no. 9, pp. 704--715, 2024.

\bibitem{zamanzadeh2024} D. Zamanzadeh \textit{et al.}, ``Data-driven prediction of continuous renal replacement therapy survival,'' \textit{Nat. Commun.}, vol. 15, p. 5440, 2024.

\bibitem{liu2022kit} L. J. Liu, V. Ortiz-Soriano, J. A. Neyra, and J. Chen, ``KIT-LSTM: Knowledge-guided time-aware LSTM for continuous clinical risk prediction,'' in \textit{Proc. IEEE BIBM}, 2022, pp. 1086--1091.

\bibitem{zha2022} D. Zha \textit{et al.}, ``Prediction model using nursing-note sentiment scores for prognosis of severe AKI patients receiving CRRT,'' \textit{Ann. Transl. Med.}, vol. 10, no. 20, p. 1110, 2022.

\bibitem{buccione2025} E. Buccione \textit{et al.}, ``Machine learning-based prediction of circuit clotting during pediatric continuous kidney replacement therapy,'' \textit{Pediatr. Nephrol.}, vol. 40, no. 12, pp. 3795--3802, 2025.

\bibitem{wang2025tmp} F. Wang \textit{et al.}, ``Real-time LSTM monitoring of dynamic transmembrane pressure to improve loop life in CRRT,'' \textit{Technol. Health Care}, vol. 33, no. 5, pp. 2305--2319, 2025.

\bibitem{vaswani2017} A. Vaswani \textit{et al.}, ``Attention is all you need,'' in \textit{Proc. NeurIPS}, 2017, pp. 5998--6008.

\bibitem{rewa2023} O. G. Rewa \textit{et al.}, ``Epidemiology and outcomes of AKI treated with continuous kidney replacement therapy: The multicenter CRRTnet study,'' \textit{Kidney Med.}, vol. 5, no. 6, p. 100641, 2023.

\bibitem{heung2017} M. Heung \textit{et al.}, ``CRRTnet: A prospective, multi-national, observational study of continuous renal replacement therapy practices,'' \textit{BMC Nephrol.}, vol. 18, no. 1, p. 222, 2017.

\bibitem{tandukar2019} S. Tandukar and P. M. Palevsky, ``Continuous renal replacement therapy: Who, when, why, and how,'' \textit{Chest}, vol. 155, no. 3, pp. 626--638, 2019.

\bibitem{macedo2016} E. Macedo and R. L. Mehta, ``Continuous dialysis therapies: Core curriculum 2016,'' \textit{Am. J. Kidney Dis.}, vol. 68, no. 4, pp. 645--657, 2016.

\bibitem{kdigo2012} Kidney Disease: Improving Global Outcomes (KDIGO) AKI Work Group, ``KDIGO clinical practice guideline for acute kidney injury,'' \textit{Kidney Int. Suppl.}, vol. 2, no. 1, pp. 1--138, 2012.

\bibitem{christ2018} M. Christ, N. Braun, J. Neuffer, and A. W. Kempa-Liehr, ``Time series feature extraction on basis of scalable hypothesis tests (tsfresh -- a Python package),'' \textit{Neurocomputing}, vol. 307, pp. 72--77, 2018.

\bibitem{welch1967} P. Welch, ``The use of fast Fourier transform for the estimation of power spectra,'' \textit{IEEE Trans. Audio Electroacoust.}, vol. 15, no. 2, pp. 70--73, 1967.

\bibitem{bandt2002} C. Bandt and B. Pompe, ``Permutation entropy: a natural complexity measure for time series,'' \textit{Phys. Rev. Lett.}, vol. 88, no. 17, p. 174102, 2002.

\bibitem{peng2005} H. Peng, F. Long, and C. Ding, ``Feature selection based on mutual information criteria of max-dependency, max-relevance, and min-redundancy,'' \textit{IEEE Trans. Pattern Anal. Mach. Intell.}, vol. 27, no. 8, pp. 1226--1238, 2005.


\bibitem{boyd2013} K. Boyd, K. H. Eng, and C. D. Page, ``Area under the precision-recall curve: point estimates and confidence intervals,'' in \textit{Proc. Joint Eur. Conf. Mach. Learn. Knowl. Discovery Databases (ECML PKDD)}, 2013, pp. 451--466.

\bibitem{chai2014} T. Chai and R. R. Draxler, ``Root mean square error (RMSE) or mean absolute error (MAE)?,'' \textit{Geosci. Model Dev. Discuss.}, vol. 7, no. 1, pp. 1525--1534, 2014.

\bibitem{wang2017tsc} Z. Wang, W. Yan, and T. Oates, ``Time series classification from scratch with deep neural networks: A strong baseline,'' in \textit{Proc. Int. Joint Conf. Neural Netw. (IJCNN)}, 2017, pp. 1578--1585.

\bibitem{cho2014gru} K. Cho \textit{et al.}, ``Learning phrase representations using RNN encoder-decoder for statistical machine translation,'' in \textit{Proc. EMNLP}, 2014, pp. 1724--1734.

\bibitem{nie2023patchtst} Y. Nie, N. H. Nguyen, P. Sinthong, and J. Kalagnanam, ``A time series is worth 64 words: Long-term forecasting with Transformers,'' in \textit{Proc. ICLR}, 2023.

\bibitem{taylor2001} G. B. Taylor Jr., C. H. Toh, W. K. Hoots, H. Wada, and M. Levi, ``Towards definition, clinical and laboratory criteria, and a scoring system for disseminated intravascular coagulation,'' \textit{Thromb. Haemost.}, vol. 86, no. 5, pp. 1327--1330, 2001.

\bibitem{evans2021} L. Evans \textit{et al.}, ``Surviving Sepsis Campaign: International guidelines for management of sepsis and septic shock 2021,'' \textit{Intensive Care Med.}, vol. 47, no. 11, pp. 1181--1247, 2021.

\bibitem{soeters2019} P. B. Soeters, R. R. Wolfe, and A. Shenkin, ``Hypoalbuminemia: Pathogenesis and clinical significance,'' \textit{JPEN J. Parenter. Enteral Nutr.}, vol. 43, no. 2, pp. 181--193, 2019.



\end{thebibliography}
\end{document}